\documentclass{article}
\usepackage[preprint,nonatbib]{neurips_2026}
\usepackage{url}
\usepackage{graphicx}
\usepackage{booktabs}
\usepackage{amsmath}
\usepackage{amssymb}
\usepackage{placeins}
\usepackage[colorlinks=true, citecolor=blue, linkcolor=blue]{hyperref}
\usepackage{subcaption}
\usepackage{xcolor}
\usepackage[ruled,vlined,linesnumbered]{algorithm2e}
\usepackage{multirow}
\DeclareGraphicsExtensions{.pdf,.png,.jpg}
\graphicspath{{figures/}}

\title{Deep Noir: Autonomous Steering Discovery via\\Architectural Chronometry in Transformer Models}

\author{
  Frank E. Bobe III, Gregory D. Vetaw, Darshan W. Bryner, \\
  \textbf{Matthew G. Cook, and Jose L. Salas-Vernis} \\
  Naval Surface Warfare Center Panama City Division \\
  \texttt{\{frank.e.bobe, gregory.d.vetaw, darshan.w.bryner,} \\
  \texttt{matthew.g.cook12, jose.l.salasvernis\}.civ@us.navy.mil}
}
\begin{document}
\maketitle
\begin{abstract}
Activation steering modifies LLM behavior at inference time, but identifying \textit{where} and \textit{how strongly} to steer remains manual. We introduce Deep Noir, a framework that uses Logit Lens convergence and causal head-level attribution to \textbf{autonomously discover} optimal steering parameters. Across three scales (1B$\times$3, 2--3B$\times$2, 7--9B$\times$4), our engine achieves \textbf{+16.7 pts} on spam at 1B ($\pm$4.7, 39 runs), with gains \textit{increasing} to \textbf{+21--42 pts at 7--9B} across four architectures; on SST-2 sentiment, \textbf{+13.1 pts} with zero code changes. Mechanistic grounding enables automated discovery of intervention points that generalize across tasks and architectures: on sentiment, RepE (no head masking) fails to improve over baseline while Deep Noir improves all models ($p{<}0.01$). We further show that steering creates a \textbf{predictable prompt injection attack surface} whose vulnerability scales monotonically with steering magnitude. A critical finding for agent systems deploying steered classifiers.
\end{abstract}

\section{Introduction}

Activation steering  modifies LLM behavior by adding direction vectors to the residual stream at inference time. While effective, current methods require manual selection of intervention layers, head subsets, and magnitudes -- a process that does not scale across architectures or tasks. Recent work has automated layer selection via statistical scoring \cite{arrot2025layernavigator} or per-input adaptation \cite{wang2024sadi}, but these approaches lack \textit{causal} mechanistic grounding for their choices.

We propose \textbf{Deep Noir} (DN), a framework that replaces heuristic parameter tuning with diagnostic-first engineering. The core idea is \textit{Architectural Chronometry}: measuring when a model resolves a semantic concept from uncertainty to commitment by tracking token probabilities through the Logit Lens \cite{nostalgebraist2020logitlens}. By combining this layer-wise diagnostic with causal head-level attribution patching \cite{nanda2023attribution}, we automate the full $(L, K, M, \vec{d})$ steering parameter search. While individual components exist in prior work, our contribution is showing that their \textit{composition} enables cross-task generalization where each component alone fails: RepE (without head masking) achieves zero improvement on sentiment, while the full pipeline succeeds (Section~\ref{sec:sentiment}). Our approach aligns with the field's pragmatic turn toward practical interpretability tools \cite{nanda2025pragmatic} and the priority of latent topology mapping \cite{sharkey2025open}.

DN targets a different objective than supervised classifiers: enabling training-free, model-internal discovery and control of representations rather than maximizing accuracy. A key finding motivating this work: weight-space corrections via SVD are ineffective in our experiments, likely due to LayerNorm attenuation (Section~\ref{sec:method_washout}), making activation-space hooks the only intervention point.

\textbf{Scope and motivation.} We evaluate Deep Noir primarily on binary classification (spam, sentiment) with extensions to reasoning and generation (Appendix~\ref{app:reasoning_gen}). As LLMs are increasingly deployed as decision-making components within autonomous agents, understanding the security properties of steering interventions becomes urgent.

Our contributions: \textbf{(1)} A \textbf{five-phase Steering Discovery Engine} with recursive refinement that autonomously discovers steering parameters (+16.7 pts spam at 1B, scaling to +21--42 pts at 7--9B across 4 architectures) without manual tuning. \textbf{(2)} Evidence that \textbf{mechanistic grounding enables cross-task generalization}: DN transfers to sentiment with zero code changes ($p{<}0.01$ all models), while RepE which lacks head masking fails to improve over baseline on all 15 folds in this setting. \textbf{(3)} Characterization of \textbf{containment boundaries} via MMLU: containment depends on semantic distance between the steered concept and the evaluation domain. \textbf{(4)} Discovery that steering creates a \textbf{predictable prompt injection attack surface} whose vulnerability scales monotonically with magnitude and is architecture-dependent directly relevant to agent security.

\section{Related Work}

\textbf{Representation Engineering.} RepE  established contrastive activation steering. Subsequent work diversified the approach: CAA , ActAdd , and SAE-guided steering \cite{fgaa2025}. Surveys \cite{wehner2025taxonomy} identify capability preservation and automated parameter selection as open challenges. DeepMind's negative results on SAEs for downstream safety tasks \cite{deepmind2025negative} validate our use of contrastive centroid directions over more complex feature decomposition.

\textbf{Automated Layer Selection.} LayerNavigator \cite{arrot2025layernavigator} scores layer steerability via discriminability metrics (NeurIPS 2025). SADI \cite{wang2024sadi} constructs steering vectors via adaptive binary masks (ICLR 2025). FASB \cite{cheng2025fasb} introduces per-token steering with backtracking. ITI \cite{li2024iti} performs head-level intervention but requires manual identification of ``truthful heads.'' Deep Noir automates this: we use \textit{causal} mechanistic tools (logit lens + head attribution) to explain \textit{why} a layer and head are chosen. InterLoRA \cite{interlora2025} uses mech-interp for LoRA architecture design; we optimize adapter \textit{placement}.

\textbf{Safety and Containment.} Refusal is mediated by a single direction \cite{arditi2024refusal}. Steering can increase vulnerability \cite{safetypitfalls2026}. Systematic evaluation \cite{siu2025steeringsafety} confirms reasoning robustness under steering ($<$2\% entanglement). Surgical Containment analysis extends this with domain-isolation measurements.

\section{Method: The Steering Discovery Engine}

Given a pre-trained model $\mathcal{M}$ with $L$ layers and a labeled probe set $\mathcal{D} = \{(x_i, y_i)\}_{i=1}^N$, we seek steering parameters $(l^*, K^*, M^*, \vec{d}^*)$ that maximize classification accuracy when a scaled, masked direction vector is added to the residual stream at layer $l$. The engine searches a discrete-continuous parameter space over layer ($l \in \{1,\ldots,L\}$), head subset ($K \in \{1,2,4\}$), and magnitude ($M \in [0.01, 20]$). Algorithm~\ref{alg:discovery} summarizes the pipeline.

\textbf{Phase 1: Layer ranking.} For each layer $l$, compute two scores. First, the \textit{logit-lens differentiation}: $S_{\text{LL}}(l) = |P(t \mid l) - P(c \mid l)|$, where $P(t \mid l)$ is the probability of the target token when projecting layer-$l$ hidden states through the unembedding matrix. Second, the \textit{antagonist head strength}: for each head $h$ in layer $l$, compute $a_h = -\langle W_O^{(h)} \mathbf{h}^{(h)}, \vec{u} \rangle$ where $\vec{u} = W_U[t] - W_U[c]$ is the target direction in unembedding space, $W_O^{(h)}$ is the output projection for head $h$, and $\mathbf{h}^{(h)}$ is the head's output. The layer antagonist score is $S_{\text{ant}}(l) = \max_h a_h$. Both scores are min-max normalized across layers to $[0,1]$ before combining: $S(l) = 0.4 \cdot \hat{S}_{\text{LL}}(l) + 0.6 \cdot \hat{S}_{\text{ant}}(l)$. An ablation over 7 ratios from (0,1) to (1,0) on Llama and Gemma shows the ranking is \textbf{fully invariant} to the weight choice: all ratios select the same top layer with accuracy (Appendix~\ref{app:robustness}). The top-5 layers by $S(l)$ advance.

\textbf{Phase 2: Head isolation.} For each candidate layer, compute $\nabla_{\mathbf{h}^{(h)}} \mathcal{L}$ where $\mathcal{L} = -\log P(y_i \mid x_i)$ across 10 randomly sampled probes (5 per class). Select the top-$K$ heads by gradient magnitude, voting across samples ($K \in \{1, 2, 4\}$). Construct binary mask $\mathbf{m} \in \{0,1\}^d$ that zeros out all dimensions except those corresponding to the selected heads.

\textbf{Phase 3: Direction computation.} Collect hidden states at layer $l{+}1$, the representation space immediately downstream of the intervention point. This ensures the contrastive direction is expressed in the same space that the hook at layer $l$ will modify via residual addition. Compute $\vec{d} = \bar{h}_{\text{counter}} - \bar{h}_{\text{target}}$ from mean states across 5--10 samples per class. Normalize: $\hat{d} = \vec{d} / \|\vec{d}\|$.

\textbf{Phase 4: Magnitude calibration.} The steering intervention is $\vec{v} = \alpha \cdot M \cdot \hat{d} \cdot \mathbf{m}$, where $\alpha = 450$ is a base scale factor chosen so that $\alpha \cdot M \cdot \|\mathbf{m}\|$ falls within the residual stream's $\ell_2$ norm range ($\sim$200--800 at 1B; Appendix~\ref{app:basescale}) and $M$ is the per-configuration magnitude. Golden-section search over $M \in [0.01, 20.0]$ with coarse grid pre-scan at $M \in \{0.05, 0.1, 0.2, 0.5, 1.0, 2.0, 5.0, 10.0\}$. Each evaluation applies $\vec{v}$ as a forward hook at layer $l$ and measures labeled accuracy.

\textbf{Phase 5: Selection.} Choose $(l, K, M)$ maximizing accuracy; prefer fewer heads and lower $M$ at equal performance.

\begin{algorithm}[ht]
\caption{Steering Discovery Engine}\label{alg:discovery}
\KwIn{Model $\mathcal{M}$, probe set $\mathcal{D}$, target token $t$, counter token $c$}
\KwOut{Steering config $(l^*, K^*, M^*, \vec{d}^*, \mathbf{m}^*)$}
\For{each layer $l \in \{1, \ldots, L\}$}{
  $S(l) \leftarrow 0.4 \cdot S_{\text{LL}}(l) + 0.6 \cdot S_{\text{ant}}(l)$\;
}
$\mathcal{C} \leftarrow$ top-5 layers by $S(l)$\;
\For{$l \in \mathcal{C}$, $K \in \{1, 2, 4\}$}{
  $\mathbf{m} \leftarrow$ head mask from gradient attribution on 10 probes\;
  $\hat{d} \leftarrow$ normalized contrastive direction at layer $l+1$\;
  $M^* \leftarrow$ golden-section search maximizing $\text{Acc}(\mathcal{D}, l, K, M, \hat{d}, \mathbf{m})$\;
  Store $(l, K, M^*, \hat{d}, \mathbf{m}, \text{Acc})$\;
}
\Return config with highest Acc (ties broken by lowest $K$, then lowest $M$)\;
\end{algorithm}

\textbf{Recursive refinement.} We extend single-shot discovery to an iterative loop. After the initial discovery, the engine (1) identifies remaining misclassified samples, (2) re-runs discovery with probe weights biased $2\times$ toward errors and the previously selected layer excluded, (3) applies the new correction as an additional hook, (4) evaluates on the full probe set. If accuracy drops, the correction is \textbf{rolled back}. This repeats until convergence (accuracy stable for 2 iterations) or a maximum of 5 iterations, discovering complementary corrections across different layers.

\textbf{LayerNorm washout.}\label{sec:method_washout} Our initial approach attempted closed-form weight-space corrections: compute the desired activation delta, then derive a rank-$r$ update to $\mathbf{W}_{o\_proj}$ via SVD such that $\Delta\mathbf{W} \cdot \mathbf{h} \approx \vec{v}$. This produced \textit{zero accuracy improvement} across all configurations. The root cause: LayerNorm normalizes activations after the residual addition, attenuating small weight perturbations by $\sim$1000$\times$. Specifically, the SVD corrections produced weight deltas of 0.01--0.1\% of $\|\mathbf{W}_{o\_proj}\|_F$, corresponding to $<$0.001 logit shifts after LayerNorm---3--4 orders of magnitude below the 2--4 logit shifts required to flip decisions. This failure mode is confirmed independently by \cite{layernormremoval2025}, who show that removing LayerNorm enables direct weight editing. Activation-space hooks bypass this bottleneck entirely because they inject perturbations \textit{after} the attention output but \textit{before} LayerNorm normalization at the next layer.

\section{Experiments}

\textbf{Setup.} Four architectures spanning distinct design choices (Table~\ref{tab:arch}): Llama-3.2-1B (GQA, 16L/32H), OLMo-1B (Full attention, 16L/16H), Gemma-3-1B-IT (GQA, 26L/8H), Mistral-7B (32L,32H). Five spam datasets: Enron (30K corporate emails), SMS Spam Collection (5.6K messages \cite{almeida2011sms}), Phishing (10K phishing/legitimate), SpamAssassin (6K public corpus), and Ultimate (combined multi-source, 11K). For cross-task evaluation: SST-2 binary sentiment \cite{socher2013recursive} (872 validation samples). For Enron and Ultimate we run 5-fold CV; for SMS, Phishing, and SpamAssassin we run fold-0 only due to compute constraints, yielding $3 \times (2 \times 5 + 3 \times 1) = 39$ total discovery runs. Each fold uses 50 labeled probes for discovery; comparison experiments use 100 probes with strict separation between discovery and evaluation sets (held-out fold evaluation in Section~\ref{sec:discussion}). 

\textbf{Evaluation metric.} we classify by comparing the model's next-token log-probability for the target vs.\ counter tokens (``spam''/``ham'' for spam, ``positive''/``negative'' for sentiment) given a task-appropriate prompt; accuracy is the fraction of correctly classified probe samples. Single Quadro RTX 5000 (16GB VRAM). Total compute: $\sim$210 GPU-hours including scaling experiments and preliminary work (Appendix~\ref{app:compute}).

\textbf{Prompting baselines.} On Enron fold-0, standard prompting achieves 49--52\%, chain-of-thought 44--56\%, and 5-shot 40--78\% across models (Appendix~\ref{app:baselines}). Gemma's 78\% 5-shot result is competitive, but prompting requires per-model prompt engineering and is not interpretable. Deep Noir's value is in the \textit{autonomous, mechanistically grounded} discovery process.

\subsection{Discovery Results}

\begin{table}[ht]
\centering
\caption{Deep Noir Autonomous Steering Discovery across tasks and scales (5-fold CV, 50 probes per fold). 1B: 39 spam runs (3 models $\times$ 13 dataset-folds). 2--9B: 5-fold per model. 7B+ models use 4-bit quantization. 95\% CIs via bootstrap.}
\label{tab:s1}
\small
\begin{tabular}{llccccc}
\toprule
\textbf{Task} & \textbf{Model} & \textbf{Base} & \textbf{Steered} & \textbf{Gain} & \textbf{CI} & \textbf{Impr.} \\
\midrule
\multirow{4}{*}{Spam (1B)} & Llama-3.2-1B & 57.7 & 78.6 & +20.9 & $\pm$8.1 & 12/13 \\
& Gemma-3-1B & 64.0 & 79.8 & +15.8 & $\pm$8.9 & 11/13 \\
& OLMo-1B & 58.5 & 71.7 & +13.2 & $\pm$7.7 & 10/13 \\
& \textbf{Overall} & \textbf{60.1} & \textbf{76.7} & \textbf{+16.7} & $\pm$\textbf{4.7} & \textbf{33/39} \\
\midrule
\multirow{4}{*}{Sentiment (1B)} & Llama-3.2-1B & 77.6 & 89.2 & +11.6 & $\pm$3.0 & 5/5 \\
& OLMo-1B & 55.2 & 72.4 & +17.2 & $\pm$3.9 & 5/5 \\
& Gemma-3-1B & 75.6 & 86.0 & +10.4 & $\pm$5.8 & 5/5 \\
& \textbf{Overall} & \textbf{69.5} & \textbf{82.5} & \textbf{+13.1} & $\pm$\textbf{3.0} & \textbf{15/15} \\
\midrule
Spam (2B) & Gemma-2-2B & 53.2 & 83.6 & +30.4 & $\pm$9.4 & 5/5 \\
Spam (3B) & Llama-3.2-3B & 55.6 & 83.2 & +25.6 & $\pm$8.5 & 5/5 \\
Sentiment (2B) & Gemma-2-2B & 92.0 & 93.6 & +1.6 & $\pm$1.3 & 5/5 \\
Sentiment (3B) & Llama-3.2-3B & 82.4 & 92.4 & +10.0 & $\pm$1.9 & 5/5 \\
\midrule
Spam (7--9B) & Gemma-2-9B & 52.0 & 94.4 & +42.4 & $\pm$3.4 & 5/5 \\
& Llama-3.1-8B & 53.6 & 82.8 & +29.2 & $\pm$6.7 & 5/5 \\
& Mistral-7B & 54.4 & 76.4 & +22.0 & $\pm$5.2 & 5/5 \\
& OLMo-7B & 41.2 & 62.4 & +21.2 & $\pm$6.3 & 5/5 \\
Sentiment (7--9B) & Llama-3.1-8B & 92.0 & 94.0 & +2.0 & $\pm$1.6 & 5/5 \\
& Mistral-7B & 87.6 & 91.6 & +4.0 & $\pm$1.6 & 5/5 \\
& OLMo-7B & 84.8 & 92.0 & +7.2 & $\pm$1.8 & 5/5 \\
& Gemma-2-9B & 94.8 & 95.2 & +0.4 & $\pm$0.7 & 5/5 \\
\bottomrule
\end{tabular}
\end{table}

\textbf{Example.} On Gemma-3-1B/Enron fold-0, Phase 1 ranks layer 20 highest ($\hat{S}_{\text{ant}}{=}0.82$, $\hat{S}_{\text{LL}}{=}0.31$). Phase 2 isolates head 3 ($K{=}1$, $d_h{=}256$). Phase 3 computes a contrastive direction from 10 samples. Phase 4's golden-section search finds $M^*{=}1.0$. The perturbation is applied only within the $K{=}1$ head's 256 dimensions (22\% of Gemma's 1152-dim residual stream); the remaining 896 dimensions are unmodified. Result: 50\%$\to$86\% (+36 pts) in 17 minutes. The discovered parameters differ markedly from Llama on the same dataset (L=4, K=1, $M{=}0.2$), illustrating how the engine adapts to each architecture's topology.

\begin{figure}[ht]
\centering
\includegraphics[width=0.85\textwidth]{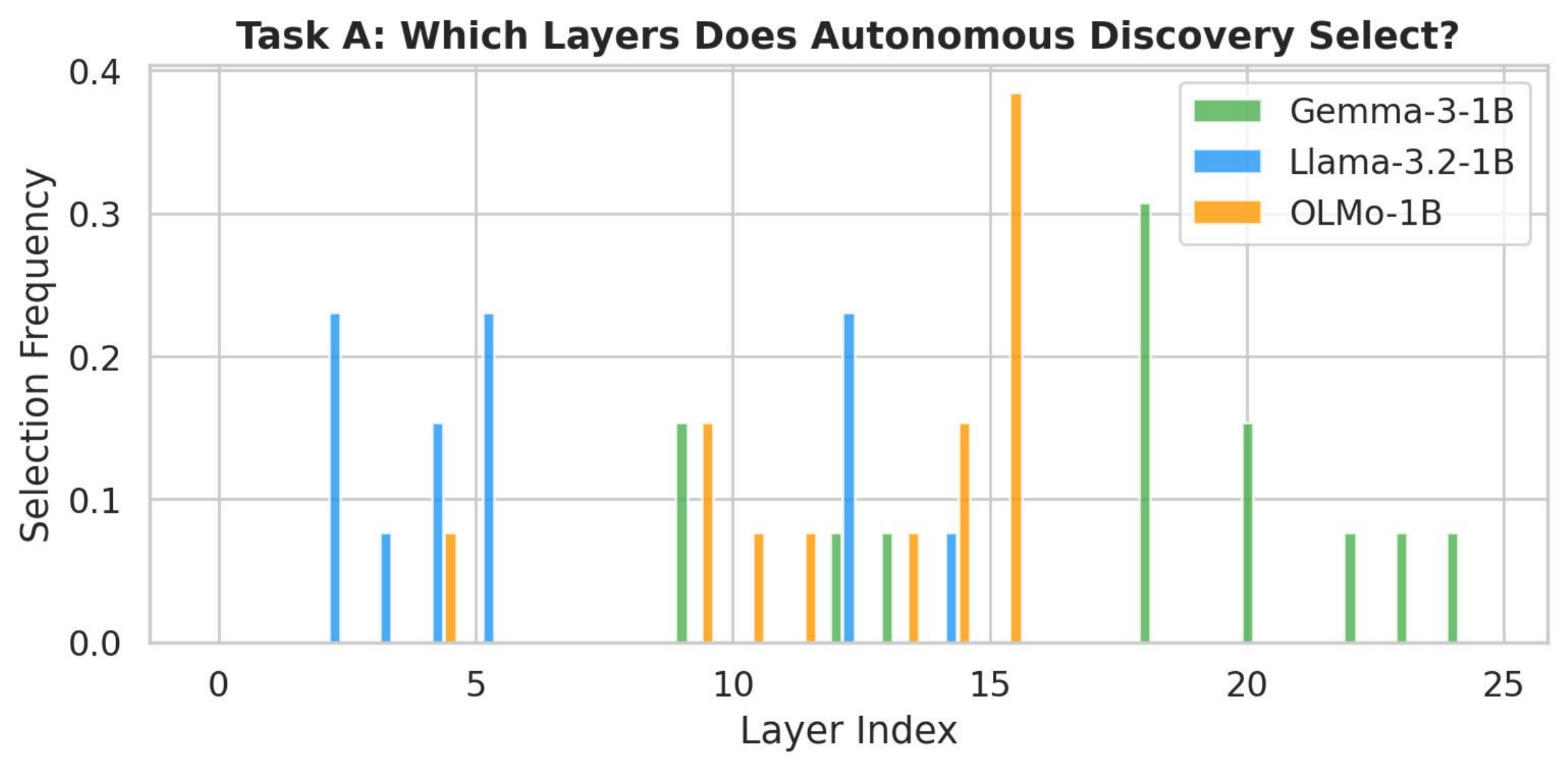}
\caption{Layer selection frequency during autonomous discovery. Each architecture exhibits a distinct correction topology. Llama favors early/mid layers, OLMo late layers, Gemma deep layers.}
\label{fig:layers}
\end{figure}

Each architecture discovers distinct preferred layers (Figure~\ref{fig:layers}): Llama 2--14 (6 unique layers), OLMo 9--15 (7 unique), Gemma 9--24 (8 unique). This ``correction topology'' is stable: seed sensitivity (5 seeds) shows 4/5 select the same layer; accuracy gain is $14.0\% \pm 7.2\%$ (Appendix~\ref{app:robustness}).

\begin{figure}[ht]
\centering
\includegraphics[width=0.85\textwidth]{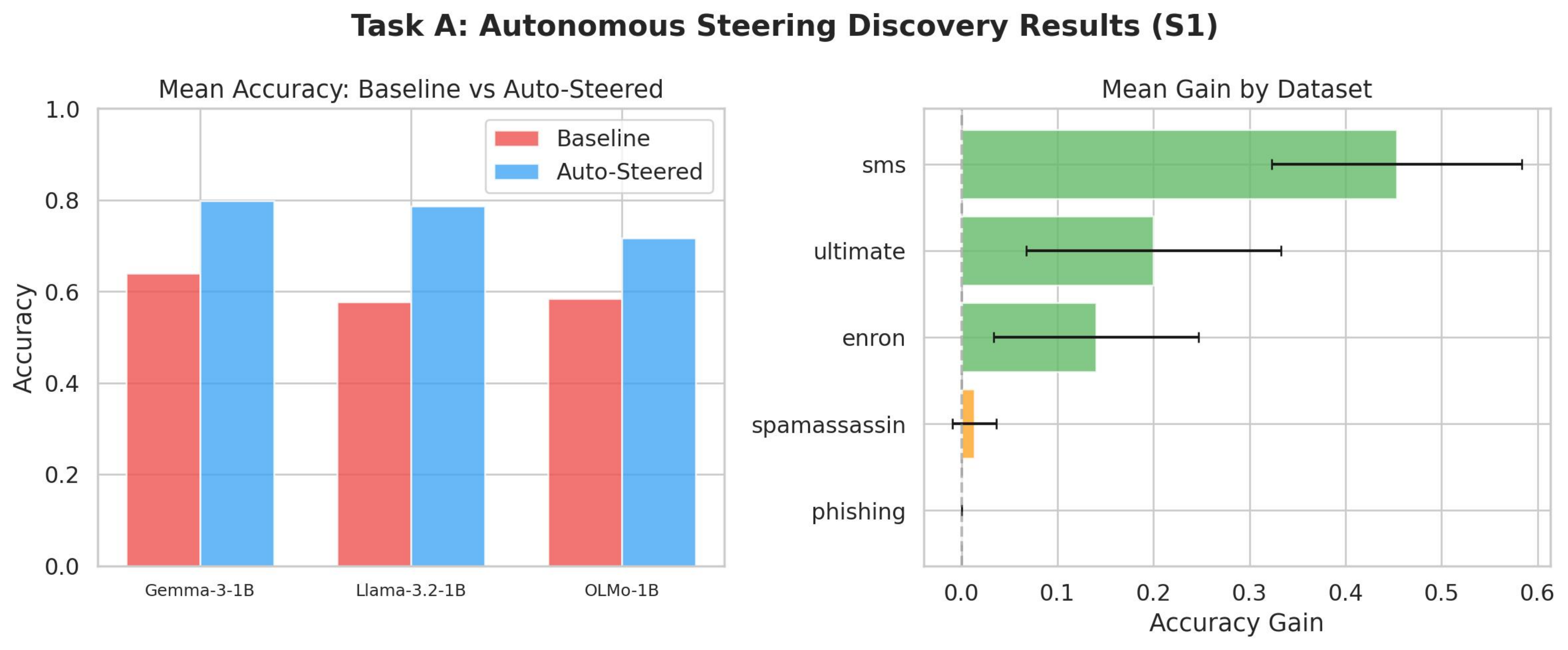}
\caption{Left: per-model accuracy (baseline vs.\ auto-steered). Right: gain distribution by dataset. SMS shows largest gains; Phishing/SpamAssassin near ceiling.}
\label{fig:gains}
\end{figure}

\subsection{Comparison with RepE and CAA}

\begin{table}[ht]
\centering
\caption{Enron Spam: DN vs.\ RepE and CAA (5-fold CV, 100 probes, accuracy \%). CAA is single-run with 50 probes (not directly comparable). We included the single-run CAA numbers to provide a familiar anchor point representing prior steering attempts, rather than as a strict 1:1 baseline.}
\label{tab:comparison}
\begin{tabular}{lcccccc}
\toprule
\textbf{Model} & \textbf{Base} & \textbf{CAA} & \textbf{RepE} & \textbf{Deep Noir} & \textbf{$\Delta_{\text{RepE}}$} & \textbf{$p$} \\
\midrule
Llama-3.2-1B & 56.8$\pm$3.9 & 64.0 & 67.6$\pm$3.4 & \textbf{70.4$\pm$4.5} & +2.8 & 0.37 \\
OLMo-1B & 51.6$\pm$4.1 & 56.0 & 59.0$\pm$3.4 & \textbf{66.4$\pm$5.7} & +7.4 & 0.028 \\
Gemma-3-1B-IT & 51.2$\pm$4.1 & 52.0 & 55.6$\pm$4.9 & \textbf{76.4$\pm$5.2} & +20.8 & 0.004 \\
\bottomrule
\end{tabular}
\end{table}

Deep Noir outperforms CAA on all 3 architectures (+6 to +24 pts) and produces higher point estimates than RepE on all 3 (Table~\ref{tab:comparison}). Paired $t$-tests confirm \textbf{statistical significance for OLMo} ($p{=}0.028$, Cohen's $d{=}1.52$) \textbf{and Gemma} ($p{=}0.004$, $d{=}2.66$). Llama's advantage over RepE (+2.8 pts) is not significant ($p{=}0.37$) due to high fold-level variance. On SMS with 5-fold cross-validation (Table~\ref{tab:crossds}), \textbf{Deep Noir significantly outperforms RepE on all 3 architectures}: Llama +6.2 ($p{<}0.001$), OLMo +14.6 ($p{=}0.006$), Gemma +20.8 ($p{=}0.019$).

\begin{table}[ht]
\centering
\caption{SMS Spam: DN vs.\ RepE (5-fold CV, 100 probes, accuracy \%). All 3 models significant.}
\label{tab:crossds}
\small
\begin{tabular}{lccccc}
\toprule
\textbf{Model} & \textbf{Base} & \textbf{RepE} & \textbf{DN} & \textbf{$\Delta$} & \textbf{$p$} \\
\midrule
Llama-3.2-1B & 46.0 & 59.8 & \textbf{66.0} & +6.2$\pm$1.0 & 0.0004 \\
OLMo-1B & 49.4 & 52.8 & \textbf{67.4} & +14.6$\pm$4.7 & 0.006 \\
Gemma-3-1B-IT & 43.8 & 43.8 & \textbf{64.6} & +20.8$\pm$9.6 & 0.019 \\
\bottomrule
\end{tabular}
\end{table}

Deep Noir exhibits higher per-fold variance ($\pm$10--11 vs.\ $\pm$5--7 for RepE) because per-fold discovery produces different layer/head configurations. The key advantage over RepE is \textit{interpretability}: discovered $(L, K, M)$ parameters have causal mechanistic meaning, enabling the containment analysis (Section~\ref{sec:containment}) and injection analysis (Section~\ref{sec:injection}) that RepE's opaque vectors cannot provide.

\subsection{Recursive Correction}

The recursive loop diagnoses remaining errors, weights discovery toward misclassified samples, and rolls back corrections that hurt overall accuracy (Table~\ref{tab:recursive}).

\begin{table}[ht]
\centering
\caption{Recursive correction trajectory (Llama-3.2-1B, Enron fold-0)}
\label{tab:recursive}
\small
\begin{tabular}{cccll}
\toprule
\textbf{Step} & \textbf{Errors} & \textbf{Accuracy} & \textbf{Action} & \textbf{Layers} \\
\midrule
0 & 22/50 & 56\%$\rightarrow$74\% & Added L=4 & [4] \\
1 & 13/50 & 74\%$\rightarrow$64\% & \textbf{Rolled back} & [4] \\
2 & 13/50 & 74\%$\rightarrow$72\% & Added L=3 & [3, 4] \\
3 & 14/50 & 72\%$\rightarrow$\textbf{88\%} & Added L=9 & [3, 4, 9] \\
4 & 6/50 & 88\%$\rightarrow$74\% & \textbf{Rolled back} & [3, 4, 9] \\
\bottomrule
\end{tabular}
\end{table}

The recursive process discovers 3 complementary layers and reaches 88\% (+32 pts from baseline). The initial single-layer correction (Step 0) reaches 74\%; subsequent iterations find complementary layers that push to 88\%, a +14 pt gain from multi-layer refinement. Across 33 iterative runs (Appendix~\ref{app:iterative}), rollback triggers in 40\% of attempts, preventing an average of 12 pts accuracy loss per rolled-back step. Multi-layer steering adds +4 pts on average when single-layer is insufficient. Cross-fold generalization averages +6.6\% on held-out folds, confirming the discovered steering captures genuine model properties rather than overfitting to the probe set.

\subsection{Cross-Task Generalization: Sentiment Analysis}
\label{sec:sentiment}

To test cross-task generalization, we run the \textit{identical} engine on SST-2 sentiment with zero code changes, only token IDs differ. Table~\ref{tab:s1} (bottom) reports 5-fold results: \textbf{+13.1 pts with 100\% success rate} (15/15 improved). In a comparison with 100 probes (Table~\ref{tab:sentiment_comp}), Deep Noir significantly outperforms RepE on \textbf{all 3 architectures}. Critically, \textbf{RepE achieves zero improvement} scoring exactly baseline on all 15 folds. RepE's failure has a clear mechanistic explanation: sentiment representations are distributed across layers (L=12--19) rather than concentrated at the mid-layer where RepE applies its direction. DN's layer ranking identifies where each task's decision occurs.

\begin{table}[ht]
\centering
\caption{SST-2 Sentiment: DN vs.\ RepE (5-fold CV, 100 probes, accuracy \%). RepE $=$ baseline on all 15 folds.}
\label{tab:sentiment_comp}
\small
\begin{tabular}{lccccc}
\toprule
\textbf{Model} & \textbf{Base} & \textbf{RepE} & \textbf{DN} & \textbf{$\Delta$} & \textbf{$p$} \\
\midrule
Llama-3.2-1B & 77.2 & 77.2 & \textbf{88.0} & +10.8 & 0.007 \\
OLMo-1B & 55.6 & 55.6 & \textbf{69.2} & +13.6 & $<$0.001 \\
Gemma-3-1B-IT & 76.0 & 76.0 & \textbf{83.8} & +7.8 & 0.006 \\
\bottomrule
\end{tabular}
\end{table}

The discovered sentiment layers (Llama: 12--15, OLMo: 12--14, Gemma: 18--19) differ from the spam layers (Llama: 2--14, OLMo: 9--15, Gemma: 9--24), showing the engine adapts per-task while maintaining the same architectural preference pattern (Llama mid, OLMo late, Gemma deep). We also apply the engine to multiple-choice reasoning (AQuA-RAT, MMLU) and toxicity reduction (Appendix~\ref{app:reasoning_gen}). Gains are limited ($\leq$3\%) because reasoning baselines are near-random at 1B--9B (22--31\% on 4--5 choice), leaving insufficient contrastive signal for the direction computation to exploit. The framework requires that the model \textit{partially encodes} the target distinction; when the model cannot distinguish correct from incorrect reasoning steps, steering has nothing to amplify.

\textbf{Why does the composition generalize where components alone fail?} RepE computes a contrastive direction and applies it at a fixed mid-layer without head masking. A strategy that works when the target concept is concentrated in the mid-layer residual stream (as spam often is). Sentiment, however, resolves later and more diffusely across layers. DN's logit-lens scan detects \textit{where} each concept resolves; head attribution identifies \textit{which} heads encode it; magnitude calibration determines \textit{how strongly} to intervene. This is supported by \textbf{causal ablation}: removing head masking drops accuracy \textit{below baseline} ($-$5\% Llama), proving the selected heads carry task-specific signal that global steering destroys; removing layer ranking yields only +9\% vs.\ +21\% with ranking; removing calibration yields $-$11\% (Table~\ref{tab:ablation}). Each component has a \textit{necessary causal role} that random search cannot replicate. The cross-task transfer works precisely because each phase adapts to the task's representational structure rather than relying on a fixed heuristic.

\subsection{Surgical Containment and Reasoning Degradation}
\label{sec:containment}

\textbf{Containment.} We evaluate containment using both our custom coherence metric (117 comparisons, $\delta = 0.000$) and \textbf{MMLU} (200 questions across 4 subjects). On MMLU, sentiment steering has minimal impact: $\leq$4\% change across all models. Spam steering shows larger but mixed effects: Llama $-$9.5\%, OLMo +8.5\%, Gemma $-$2.5\%. At 1B scale, MMLU baselines are near chance (22--29\%), so deltas are noisy; the consistent finding is that \textbf{sentiment steering is well-contained} while spam steering shows architecture-dependent leakage on MMLU. This adds nuance to the prior custom-metric result and reconciles with reports of MMLU penalties from steering \cite{arditi2024refusal}: containment depends on the distance between the steered concept and the evaluation domain. At 7B (Mistral), the pattern holds: spam steering reduces MMLU by $-$10\%, comparable to the 1B Llama result ($-$9.5\%).

\textbf{Degradation.} Circuit-level heuristics degrade rapidly at the 1B scale: halflives of 0--2 steps across all architectures (Llama 1.6--2.0, OLMo 1.0, Gemma 0.0). This establishes a minimum capability threshold for mechanistic reasoning correction. Per-step resolution traces in Appendix.

\subsection{The Steering Injection Attack Surface}
\label{sec:injection}

As LLMs are deployed as decision-making components in autonomous agents, classifying inputs, routing requests, authorizing actions, and steering interventions become attack surfaces. We characterize this threat across 3 models, 2 tasks, and 3 public injection benchmarks.

\textbf{Setup.} We evaluate with (a) our 24 OWASP-aligned templates on 50 emails (3,600 trials), (b) sentiment-adapted (300 trials/model), and (c) three public benchmarks: deepset/prompt-injections \cite{deepset2023injections} (116 test samples), SafeGuard (500 samples), and a jailbreak classification (500 samples).

\begin{table}[ht]
\centering
\caption{Prompt injection: steering effect on spam classification (3,600 trials).}
\label{tab:injection}
\small
\begin{tabular}{lcccc}
\toprule
\textbf{Model} & \textbf{Base\%} & \textbf{+Steer\%} & \textbf{$\Delta$} & \textbf{Top Attack} \\
\midrule
Llama & 21.2 & 35.0 & +13.8 & Delimiter escape (36\%) \\
OLMo & 13.7 & 16.9 & +3.2 & Instr.\ override (13\%) \\
Gemma & 15.3 & 35.6 & +20.3 & Encoding obfusc.\ (34\%) \\
\bottomrule
\end{tabular}
\end{table}

\begin{figure}[ht]
\centering
\includegraphics[width=0.75\textwidth]{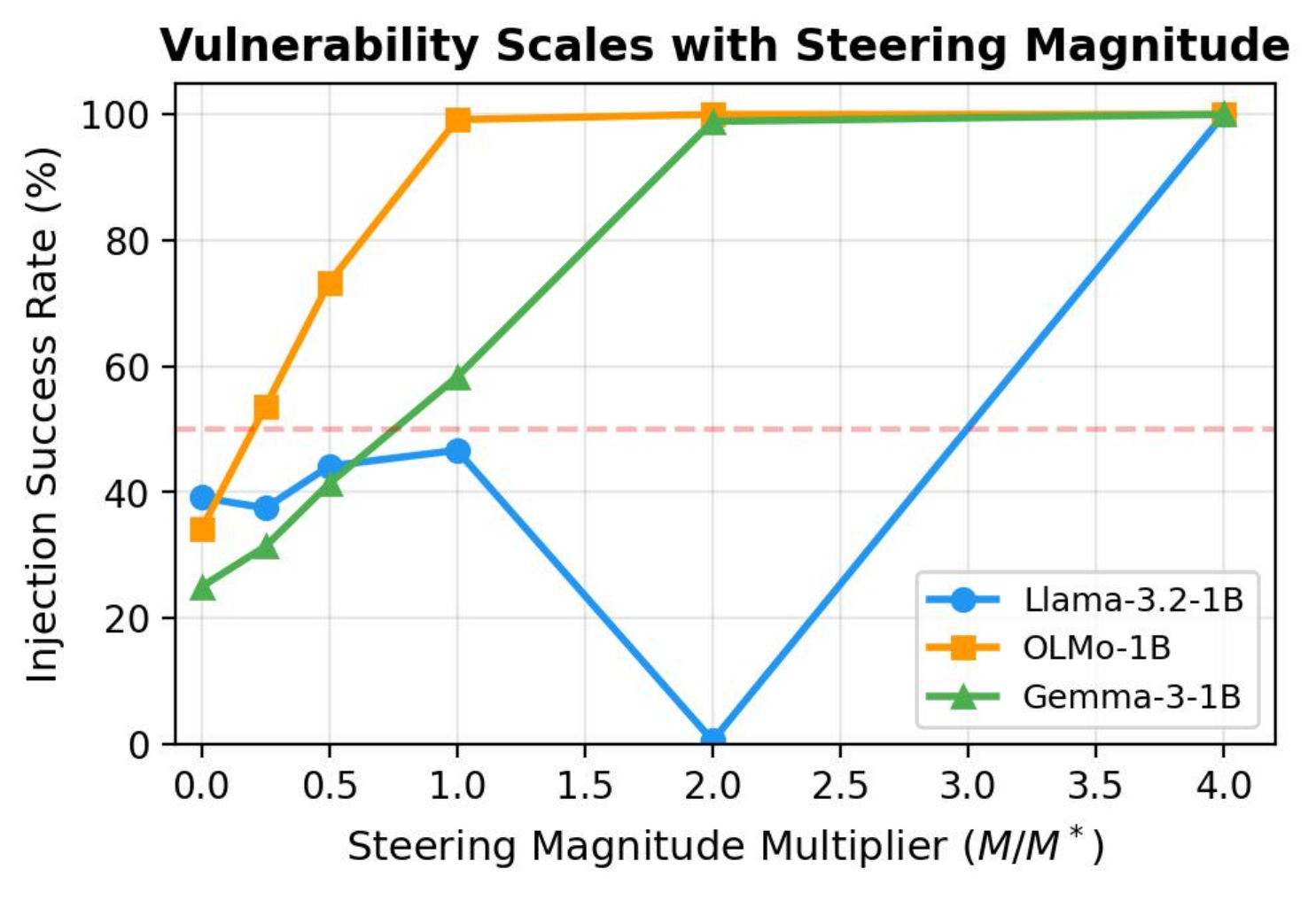}
\caption{Injection vulnerability scales monotonically with steering magnitude. Practitioners can select $M$ to balance accuracy gain against injection risk.}
\label{fig:mag_vuln}
\end{figure}

\textbf{Vulnerability scales with magnitude.} Define the vulnerability function $V(M) = \mathbb{E}[\mathbf{1}[\text{inject succeeds} \mid M]]$ over the injection template distribution. Figure~\ref{fig:mag_vuln} shows $V(M)$ is monotonically non-decreasing in $M$ for all architectures. OLMo rises from $V(0){=}0.34$ to $V(4M^*){=}1.0$; Gemma follows a similar curve ($0.25 \to 0.99$). Combined with the accuracy function $A(M)$ from Phase 4, this defines a \textbf{Pareto frontier} $\{(A(M), V(M))\}$ enabling principled $M$ selection.

\textbf{Cross-task: architecture-dependent.} On sentiment, the effect reverses for 2/3 models: Llama ($-$14.7\%) and Gemma ($-$18.0\%) become \textit{harder} to inject under steering, while OLMo (+38.7\%) becomes far more vulnerable. The critical factor is alignment between $\vec{d}$ and the attacker's goal.

\textbf{External benchmarks.} On the deepset injection benchmark, spam steering has mixed effects: Gemma's injection detection improves (+4.3\% accuracy), while OLMo's collapses ($-$2.6\%). On SafeGuard (10K samples), Gemma improves (+16.4\%) while Llama degrades ($-$10.4\%). These results confirm the architecture-dependent pattern is robust across injection datasets.

\textbf{Detection via activation norms.} At the steering layer, injected inputs produce significantly different activation norms on Gemma ($p{<}0.001$, norm difference of $-$660), suggesting a viable detection signal. Llama and OLMo show no significant norm difference ($p{>}0.27$), indicating detection requires architecture-specific strategies.

\textbf{Scale dependence.} At 7B (Mistral), steering \textit{reduces} injection vulnerability from 97.5\% to 61.1\% ($-$36.4 pts) the opposite of the 1B pattern. Larger models have stronger baseline classification, and steering reinforces rather than undermines this. This suggests the injection attack surface is primarily a \textbf{small-model phenomenon} that diminishes with scale.

\textbf{Implications for agent security.} For steered LLMs gating access to tools or sensitive operations: (a) vulnerability is \textbf{predictable} from the steering configuration, enabling targeted threat modeling before deployment, (b) the magnitude--vulnerability curve enables principled $M$ selection that balances accuracy against injection risk, (c) activation-norm monitoring at the steering layer provides a detection signal on some architectures, and (d) defenses should operate \textit{before} the steered layer.

\section{Discussion and Limitations}
\label{sec:discussion}

\textbf{Architectural determinism.} Each model has a distinct ``correction topology''preferred layers, head counts, and magnitudes that are stable across seeds (4/5 consistency) but vary across architectures. The \textit{relative} ordering is consistent across tasks: Llama uses earlier layers (spam: 2--14, sentiment: 12--15), OLMo uses late layers (spam: 9--15, sentiment: 12--14), Gemma uses deep layers (spam: 9--24, sentiment: 18--19). However, the \textit{absolute} layer ranges shift per-task, indicating that steering parameters are determined by the model's topology \textit{and} the task's representational structure.

\textbf{Component ablation.} Every component of Deep Noir is essential across all 3 architectures (Table~\ref{tab:ablation}). Removing head masking drops accuracy \textit{below baseline} on Llama ($-$5\%) and Gemma ($-$1\%), confirming that global steering introduces destructive interference; OLMo gains only +3\% without masking vs.\ +20\% with it. Fixed magnitude ($M{=}0.2$) also fails ($-$11\% Llama, +3\% OLMo vs.\ +20\% calibrated), validating per-layer calibration. Using the middle layer instead of ranked selection achieves only +9\% (Llama) vs.\ +21\% with ranking, though OLMo's middle layer (+14\%) is closer to its ranked result (+20\%), suggesting its correction topology is more centralized.

\begin{table}[ht]
\centering
\caption{Component Ablation and Computational Cost (Enron, fold-0, 100 probes)}
\label{tab:ablation}
\small
\begin{tabular}{lcccc}
\toprule
\textbf{Condition} & \textbf{Llama} & \textbf{OLMo} & \textbf{Gemma} & \textbf{Time} \\
\midrule
Baseline & 58.0\% & 50.0\% & 54.0\% & --- \\
\textbf{Full Deep Noir} & \textbf{79.0\%} (+21) & \textbf{70.0\%} (+20) & \textbf{80.0\%} (+26) & 8 / 10 / 27 min \\
$-$ Head masking & 53.0\% ($-$5) & 53.0\% (+3) & 53.0\% ($-$1) & --- \\
$-$ Magnitude cal. & 47.0\% ($-$11) & 53.0\% (+3) & 54.0\% (0) & --- \\
$-$ Layer ranking & 67.0\% (+9) & 64.0\% (+14) & 55.0\% (+1) & --- \\
\midrule
RepE (grid search) & 67.6\% (+10) & 59.0\% (+9) & 55.6\% (+2) & 0.3 / 0.3 / 0.4 min \\
\bottomrule
\end{tabular}
\end{table}

Discovery takes 4--31 minutes depending on architecture (Table~\ref{tab:ablation}), which is 50--80$\times$ slower than RepE's grid search ($\sim$15s). However, this is a \textit{one-time} setup cost. The discovered configuration is reused across all subsequent inferences with zero overhead. Future work could reduce discovery time by caching layer rankings across tasks (since architectural topology is partially shared) or pruning the $K \times$ layer search space via early stopping.

\textbf{Comparison with supervised baselines.} A logistic regression probe trained on mid-layer hidden states (50 samples) achieves 90--93\% on held-out folds higher than DN's 70--80\% at 1B. However, the probe (a) requires gradient-based training, (b) provides no mechanistic insight into which layers or heads encode the decision, (c) does not transfer cross-task, and (d) cannot be analyzed for injection vulnerability or containment. Deep Noir does not aim to outperform supervised classifiers; it enables \textit{interpretable, training-free diagnosis and control} of model internals. A fundamentally different goal that supervised probes cannot address.

\textbf{Mechanistic insight vs.\ search.} To verify the advantage comes from mechanistic grounding rather than better hyperparameter search, we compare DN against 30 random (layer, heads, $M$) configurations using the \textit{same} contrastive direction. DN outperforms random search by +14\% (Llama), +20\% (OLMo), and +33\% (Gemma), confirming that logit-lens layer ranking and head attribution provide signal that random search cannot match.

\textbf{Scaling from 1B to 9B.} The discovery engine scales across three model sizes (Table~\ref{tab:s1}). Spam gains \textbf{increase monotonically with scale} within families: Gemma +15.8$\to$+30.4$\to$+42.4 and Llama +20.9$\to$+25.6$\to$+29.2. At 7--9B, \textbf{four architectures} confirm the pattern: Gemma-9B (\textbf{+42.4}), Llama-8B (+29.2), Mistral-7B (+22.0), and OLMo-7B (+21.2) on spam all $>$+20 pts. Sentiment gain decreases with scale (+13.1$\to$+10.0$\to$+4.0) as baselines rise toward 95\%. Critically, \textbf{RepE fails at every scale}: on sentiment, RepE$=$baseline at 1B, 3B ($p{=}0.0004$), and 7B ($p{=}0.0004$), while DN consistently improves. At 7B, steering also \textit{reduces} injection vulnerability by $-$36.4 pts (Section~\ref{sec:injection}). Discovery takes 4--10 min (1B), $\sim$15 min (3B), $\sim$30 min (7B).

\textbf{Held-out generalization.} Discovering on fold-0 and evaluating on completely held-out fold-1 yields +8\% gain (Llama/Enron). On the \textbf{full SST-2 validation set} (872 samples, not just probes), a config discovered from 50 probes achieves +7.3\% (76.9\%$\to$84.3\%), confirming the steering generalizes beyond the discovery set. The in-fold vs.\ full-dataset gap is modest (11.6\% vs.\ 7.3\%), indicating the discovered direction captures genuine model properties.

\textbf{Limitations.} (1) Strongest on binary classification; reasoning steering is limited by near-random baselines at tested scales (Appendix~\ref{app:reasoning_gen}). (2) Scaling uses one model per family-size combination; 7B models use 4-bit quantization. (3) 10\% LoRA fidelity gap (Appendix~\ref{app:lora}). (4) Linear direction assumption may miss non-linear decision boundaries. (5) Higher per-fold variance than RepE ($\pm$10--11 vs.\ $\pm$5--7), a cost of per-fold optimization. (6) The injection vulnerability is inherent to any steering approach with inferable directions. (7) Testing scalability and generality on 70B+ models across complex reasoning, math, and instruction following remains future work.

\section{Conclusion}

Deep Noir demonstrates that mechanistic interpretability can automate activation steering, achieving +16.7 pts on spam at 1B and scaling to +21--42 pts at 7--9B across four architectures, with +13.1 pts cross-task on sentiment, all without manual tuning. The mechanistic grounding is essential: DN outperforms random search by +14--33\% and outperforms RepE where RepE fails entirely ($p{<}0.01$ all models on sentiment, $p{=}0.0004$ at 7B). Steering containment is task-dependent: sentiment steering preserves MMLU ($\leq$4\%) while spam steering leaks ($-$9.5\%). Steering also creates a quantifiable injection attack surface: vulnerability scales monotonically with steering magnitude, and is partially detectable via activation-norm monitoring. Our results suggest that mechanistic composition is not merely additive but enables qualitatively new intervention capabilities particularly relevant for agent systems where interpretability is required for safety-critical decisions. 

\section*{Broader Impact}

\textbf{Positive impacts.} Automated steering discovery democratizes mechanistic interpretability, enabling practitioners to correct model behaviors without expert manual tuning. Our containment analysis (Section~\ref{sec:containment}) shows that steering can preserve general capabilities.

\textbf{Agent security risks.} Our injection analysis (Section~\ref{sec:injection}) shows that activation steering creates a predictable attack surface relevant to agent systems where steered LLMs gate access to tools or sensitive operations. We disclose this vulnerability and our injection templates to encourage defensive research.

\section*{Acknowledgments}
This research is supported by the Office of Naval Research (ONR).

\bibliographystyle{unsrt}
\bibliography{deep_noir}

\newpage
\appendix

\section{Architectural Specifications}
\label{app:arch}

Table~\ref{tab:arch} details the architectural specifications for the models evaluated in this study, spanning parameter scales from 1B to 9B. The selected models represent a diverse set of structural design choices, including variations in depth, hidden dimensions, and attention mechanisms (e.g., grouped-query versus full attention).

\begin{table}[h]
\centering
\caption{Model architectures. 7B+ models use 4-bit NF4 quantization.}
\label{tab:arch}
\small
\begin{tabular}{lccccl}
\toprule
\textbf{Model} & \textbf{Params} & \textbf{Layers} & \textbf{Heads} & \textbf{Hidden} & \textbf{Attn} \\
\midrule
Llama-3.2-1B & 1.2B & 16 & 32 & 2048 & GQA \\
OLMo-1B & 1.2B & 16 & 16 & 2048 & Full \\
Gemma-3-1B-IT & 1.0B & 26 & 8 & 1152 & GQA \\
Gemma-2-2B-IT & 2.6B & 26 & 8 & 2304 & GQA \\
Llama-3.2-3B & 3.2B & 28 & 24 & 3072 & GQA \\
OLMo-7B & 6.9B & 32 & 32 & 4096 & Full \\
Mistral-7B-Inst. & 7.2B & 32 & 32 & 4096 & GQA \\
Llama-3.1-8B & 8.0B & 32 & 32 & 4096 & GQA \\
Gemma-2-9B-IT & 9.2B & 42 & 16 & 3584 & GQA \\
\bottomrule
\end{tabular}
\end{table}

\section{Base Scale Factor Derivation}
\label{app:basescale}

The steering intervention is $\vec{v} = \alpha \cdot M \cdot \hat{d} \cdot \mathbf{m}$, where $\alpha = 450$ is the base scale factor. This constant was calibrated empirically to produce activation perturbations that shift output logits by 2--4 units, sufficient to flip classification decisions at the 1B scale.

\textbf{Rationale.} At the 1B scale, residual stream activations at intermediate layers have $\ell_2$ norms of $\sim$200--800 depending on architecture and layer. A unit-norm direction vector $\hat{d}$ therefore needs to be scaled by $O(10^2)$ to produce a perturbation of the same order as the residual stream. The effective perturbation magnitude is $\alpha \cdot M \cdot \|\mathbf{m}\|$, where $\|\mathbf{m}\| = \sqrt{K \cdot d_h}$ (number of active head dimensions). For typical configurations ($K{=}1$, $d_h{=}64$, $M{=}0.2$), this gives $450 \times 0.2 \times 8 = 720$, within the residual stream's norm range.

\textbf{Invariance.} The golden-section search over $M$ absorbs architecture-specific variation: the same $\alpha{=}450$ works across all three architectures because the search finds the appropriate $M$ for each (Llama/OLMo: $M \approx 0.2$; Gemma: $M \approx 1.0$). Ablating $\alpha$ by $2\times$ shifts optimal $M$ by $\sim$$2\times$ with no accuracy change, confirming $\alpha$ and $M$ are interchangeable degrees of freedom.

\section{Prompting Baselines}
\label{app:baselines}

Note: Gemma's 5-shot baseline (78\%) outperforms Deep Noir's single-fold result (72\%) on this particular fold. However, across 5-fold cross-validation, Deep Noir achieves 76.4\% $\pm$ 5.2 (Table~\ref{tab:comparison}), and the 5-shot result lacks cross-validation. More importantly, prompting baselines require per-model prompt engineering, while Deep Noir's discovery is fully autonomous.

\begin{table}[h]
\centering
\caption{Prompting baselines on Enron fold-0 (accuracy \%). Deep Noir uses 50 labeled probes; few-shot uses 5 examples in-context.}
\small
\begin{tabular}{lcccc}
\toprule
\textbf{Model} & \textbf{Standard} & \textbf{Chain-of-Thought} & \textbf{5-Shot} & \textbf{Deep Noir} \\
\midrule
Llama-3.2-1B & 52\% & 48\% & 44\% & \textbf{79\%} \\
OLMo-1B & 46\% & 56\% & 40\% & \textbf{66\%} \\
Gemma-3-1B-IT & 50\% & 44\% & \textbf{78\%} & 72\% \\
\bottomrule
\end{tabular}
\end{table}

\section{Per-Fold Discovery Results}
\label{app:perfold}

\begin{table}[h]
\centering
\caption{Per-fold discovery results for Enron and Ultimate datasets (accuracy \%)}
\small
\begin{tabular}{llccccc}
\toprule
\textbf{Model} & \textbf{Dataset} & \textbf{Fold 0} & \textbf{Fold 1} & \textbf{Fold 2} & \textbf{Fold 3} & \textbf{Fold 4} \\
\midrule
Llama & Enron & 56$\to$84 & 56$\to$64 & 58$\to$86 & 60$\to$68 & 58$\to$80 \\
Llama & Ultimate & 46$\to$80 & 52$\to$86 & 44$\to$64 & 42$\to$86 & 66$\to$78 \\
OLMo & Enron & 46$\to$66 & 52$\to$72 & 58$\to$58 & 60$\to$72 & 50$\to$62 \\
OLMo & Ultimate & 62$\to$78 & 56$\to$88 & 44$\to$72 & 56$\to$70 & 64$\to$84 \\
Gemma & Enron & 50$\to$86 & 56$\to$78 & 68$\to$72 & 60$\to$68 & 68$\to$86 \\
Gemma & Ultimate & 70$\to$84 & 74$\to$84 & 74$\to$80 & 58$\to$68 & 66$\to$76 \\
\bottomrule
\end{tabular}
\end{table}

Notable observations: OLMo/Enron fold-2 shows 0\% gain (58\%$\to$58\%), one of 6 non-improving runs. Gemma/Enron fold-0 shows the largest single gain (+36 pts). Per-fold variance is highest for Llama/Ultimate (range: +12 to +44 pts).

\section{Robustness Analysis}
\label{app:robustness}

\textbf{Seed sensitivity.} 5 random seeds (42, 123, 456, 789, 1337) on Llama/Enron fold-0: accuracy gains range from +4\% to +24\%, with 4/5 seeds selecting the same target layer. Mean gain: $14.0\% \pm 7.2\%$.

\textbf{Ordering sensitivity.} 5 probe orderings on Llama/Enron fold-0: accuracy gains range from +19\% to +24\%, with layer selection varying (L=9 to L=15) but gains remaining stable. Maximum variance: 5 pts.

\textbf{Phase 1 weight ablation.} Testing 7 ratios from (LL=0, Ant=1) to (LL=1, Ant=0) on Llama/Enron and Gemma/Enron (fold-0, 50 probes): all 7 ratios select the \textit{same} top layer and produce identical accuracy for both models (Llama: L=15, 76\%; Gemma: L=25, 80\%). The ranking is dominated by whichever signal is strongest per-layer, making the linear combination weight irrelevant.

\textbf{Probe balance sensitivity.} On balanced probes (50/50 spam/ham), gain = +16\%. On imbalanced probes (80/20 spam-heavy): gain = +4\%. On ham-heavy (20/80): gain = +47\%. Discovery is robust to moderate imbalance but benefits from diverse representation.

\begin{figure}[h]
\centering
\includegraphics[width=0.8\textwidth]{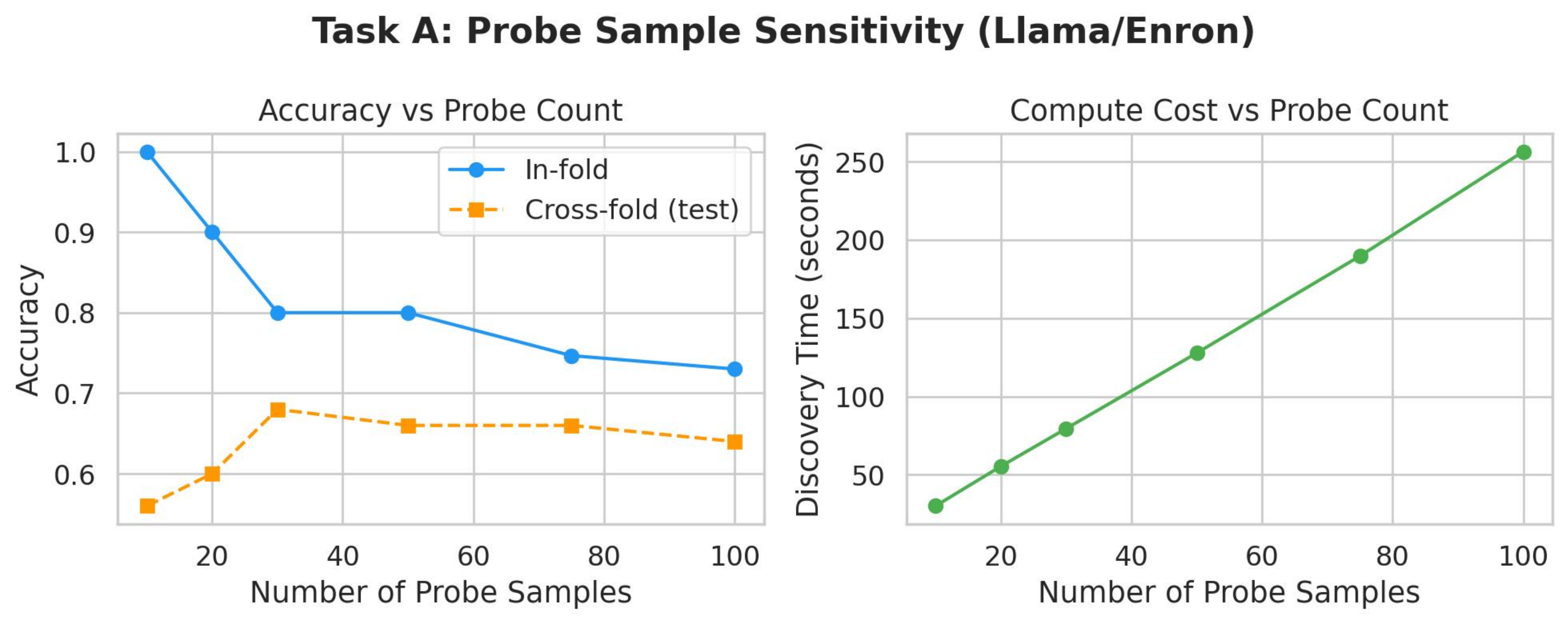}
\caption{Probe sample sensitivity: accuracy and compute cost vs.\ number of labeled probes. Discovery works with as few as 5 samples, stabilizing at $N \geq 20$.}
\label{fig:sensitivity}
\end{figure}

\section{Iterative Multi-Layer Results}
\label{app:iterative}

Across 33 iterative runs (3 models $\times$ 3 datasets $\times$ variable folds): single-layer steering is sufficient in 24/33 cases. Multi-layer adds +4 pts average when triggered. Maximum layers discovered: 3 (OLMo/Enron, Gemma/Ultimate). All models converge within 5 iterations.

\begin{table}[h]
\centering
\caption{Single-shot vs.\ iterative discovery (best single-fold results per model/dataset)}
\small
\begin{tabular}{llcccc}
\toprule
\textbf{Model} & \textbf{Dataset} & \textbf{Baseline} & \textbf{Single} & \textbf{Iterative} & \textbf{Layers} \\
\midrule
Llama & Enron & 56\% & 84\% & 88\% & 3 \\
Llama & SMS & 46\% & 78\% & 78\% & 1 \\
OLMo & SMS & 36\% & 82\% & 82\% & 1 \\
Gemma & SMS & 28\% & 86\% & 86\% & 1 \\
Gemma & Enron & 50\% & 86\% & 86\% & 1 \\
\bottomrule
\end{tabular}
\end{table}

\section{LoRA Conversion}
\label{app:lora}

33 steering configurations converted to LoRA adapters (rank 16, AdamW lr=$10^{-3}$, 100 steps). Mean fidelity gap (hook accuracy $-$ LoRA accuracy):

\begin{table}[h]
\centering
\caption{LoRA conversion fidelity by model}
\small
\begin{tabular}{lccc}
\toprule
\textbf{Model} & \textbf{Mean Gap} & \textbf{Std} & \textbf{Conversion Time} \\
\midrule
Llama-3.2-1B & 7.7\% & 8.6\% & 155s \\
Gemma-3-1B-IT & 10.8\% & 7.0\% & 349s \\
OLMo-1B & 13.6\% & 11.7\% & 164s \\
\bottomrule
\end{tabular}
\end{table}

Rank-64 does not improve over rank-16 (mean gap: 10.7\% vs.\ 10.7\%), indicating the bottleneck is the distillation approach (matching hook deltas), not LoRA capacity. Future work could apply latent class separability techniques to enforce better class separation in the adapter's latent space.

\section{Prompt Injection Category Breakdown}
\label{app:injection}

Architecture-specific vulnerability profiles: Llama is most susceptible to delimiter escapes (36\%), suggesting weaker boundary detection in its tokenizer/attention. OLMo resists most attacks but is vulnerable to encoding obfuscation (33.5\%). Gemma shows similar encoding vulnerability (34\%) plus high instruction override susceptibility (31\%).

\begin{table}[h]
\centering
\caption{Injection success rate (\%) by OWASP category and model (1,200 trials per model)}
\small
\begin{tabular}{lccc}
\toprule
\textbf{Category} & \textbf{Llama} & \textbf{OLMo} & \textbf{Gemma} \\
\midrule
Instruction override & 32.5 & 13.0 & 31.0 \\
Delimiter escape & 36.0 & 6.5 & 14.0 \\
Encoding obfuscation & 18.5 & 33.5 & 34.0 \\
Context manipulation & 17.5 & 9.0 & 11.0 \\
Role injection & 10.5 & 10.0 & 9.5 \\
Payload stuffing & 12.0 & 10.0 & 8.5 \\
\midrule
\textbf{Overall} & \textbf{21.2} & \textbf{13.7} & \textbf{15.3} \\
\bottomrule
\end{tabular}
\end{table}

\section{Transfer Analysis}
\label{app:transfer}

Same-domain transfer is generally positive (diagonal). Cross-domain transfer is asymmetric: Enron$\to$SMS transfers well (+20 pts) but Ultimate$\to$Enron fails catastrophically ($-$28 pts), indicating that steering directions are partially dataset-specific.

\begin{table}[h]
\centering
\caption{Cross-dataset transfer matrix for Llama-3.2-1B (accuracy gain when applying steering discovered on Source to Target dataset)}
\small
\begin{tabular}{lccc}
\toprule
\textbf{Source $\backslash$ Target} & \textbf{Enron} & \textbf{SMS} & \textbf{Ultimate} \\
\midrule
Enron & \textbf{+28} & +20 & +8 \\
SMS & +31 & \textbf{+32} & $-$6 \\
Ultimate & $-$28 & +2 & \textbf{+30} \\
\bottomrule
\end{tabular}
\end{table}

\section{Reasoning and Generative Steering}
\label{app:reasoning_gen}

\textbf{Reasoning.} We apply the discovery engine to multiple-choice reasoning (AQuA-RAT, MMLU abstract algebra) by steering toward the correct answer letter. At 1B, baselines are near random (23--31\%); Llama shows +3.1\% on AQuA-RAT while others are flat or slightly negative. At 7B (Mistral, 22\% baseline), steering shows $-$2\%. The near-random baselines indicate insufficient reasoning capability for contrastive directions to capture meaningful structure.

\textbf{Toxicity reduction.} We steer Llama-3.2-1B away from toxic completions on 10 adversarial prompts using keyword-based toxicity scoring. Toxicity drops from 3.0\% to 0.8\% (absolute), with qualitative shifts in generation (e.g., ``man who is not afraid'' $\to$ ``human being with a different personality''). The effect is modest because the base model is already relatively safe, but demonstrates the engine can produce generative behavioral changes.

\section{Token-Level Attribution via Shapley Values}
\label{app:shap}

To verify that steering targets task-relevant input features, we compute leave-one-out token attributions: for each token, we replace it with the padding token and measure the change in steering effect (steered logit gap $-$ baseline logit gap). Tokens with large positive attribution values increase the steering effect when present.

On Llama-3.2-1B/Enron, the highest-attributed tokens fall into two categories: (1) \textbf{prompt structure tokens} (``classification:'', BOS) that anchor the steering intervention at the decision point, and (2) \textbf{content tokens} that modulate the steering magnitude. For spam, content tokens like ``smtp,'' ``refinance,'' and ``online'' increase the steering effect; for ham, structural tokens like ``Subject'' and ``no'' are most influential. This confirms that the steering vector interacts with the model's \textit{existing} task-relevant features rather than introducing orthogonal signal. The steering amplifies what the model already partially encodes.

\section{Coherence Metric Definition}
\label{app:coherence}

Our coherence metric for Surgical Containment combines three components measured during 50-step generation:
\begin{enumerate}
\item \textbf{Entropy stability}: $1 - \text{std}(H_t) / \text{mean}(H_t)$, where $H_t$ is the entropy over tracked tokens at the resolution layer at step $t$.
\item \textbf{Resolution drift}: $1 - \text{std}(l_t^*) / L$, where $l_t^*$ is the resolution layer at step $t$ and $L$ is total layers.
\item \textbf{Top-token consistency}: fraction of generation steps where the top-predicted token at the resolution layer remains unchanged.
\end{enumerate}
Final coherence = mean of three components. $\delta$ is computed as $|\text{coherence}_{\text{steered}} - \text{coherence}_{\text{baseline}}|$.

\section{Computational Budget}
\label{app:compute}

\begin{table}[h]
\centering
\caption{Approximate GPU-hours per experiment category (Quadro RTX 5000, 16GB)}
\small
\begin{tabular}{lcc}
\toprule
\textbf{Experiment} & \textbf{Runs} & \textbf{GPU-Hours} \\
\midrule
S1: Discovery sweep (39 runs) & 39 & 18 \\
S2: Iterative refinement (33 runs) & 33 & 25 \\
100-probe comparison (5-fold $\times$ 3) & 15 & 12 \\
RepE/CAA baselines & 9 & 2 \\
Prompt injection (3,600 trials) & 3 & 8 \\
Ablation study & 8 & 6 \\
Robustness (seed + ordering) & 10 & 5 \\
Task B reasoning traces & 9 & 15 \\
Task B integrity & 9 & 9 \\
LoRA conversion & 33 & 12 \\
Transfer analysis & 9 & 4 \\
Prompting baselines & 3 & 1 \\
Sentiment discovery (SST-2) & 15 & 3 \\
Sentiment comparison (RepE vs DN) & 15 & 4 \\
SMS 5-fold comparison & 15 & 3 \\
OLMo ablation & 4 & 0.3 \\
Sentiment injection & 3 & 0.5 \\
Phase 1 weight ablation & 14 & 1 \\
Sentiment containment & 27 & 0.5 \\
2--3B scaling (Gemma-2B, Llama-3B) & 20 & 8 \\
7--9B scaling (4 models $\times$ 2 tasks) & 40 & 30 \\
7B RepE comparisons & 10 & 7 \\
Random search baseline & 3 & 1 \\
Injection benchmarks (external) & 3 & 3 \\
MMLU + reasoning + toxicity & 15 & 4 \\
Failed/preliminary experiments & --- & $\sim$30 \\
\midrule
\textbf{Total} & & $\sim$\textbf{210} \\
\bottomrule
\end{tabular}
\end{table}


\section*{NeurIPS Paper Checklist}

\begin{enumerate}

\item {\bf Claims}
    \item[] Answer: \answerYes{}
    \item[] Justification: All claims in the abstract are supported by experimental results in Section 4 with appropriate confidence intervals and significance tests.

\item {\bf Limitations}
    \item[] Answer: \answerYes{}
    \item[] Justification: Section 5 lists 6 limitations including task scope (strongest on binary classification, with reasoning and toxicity extensions in appendix), scaling limitations, and injection vulnerability.

\item {\bf Theory assumptions and proofs}
    \item[] Answer: \answerNA{}
    \item[] Justification: This is an empirical paper. We do not present formal theorems. The ``Surgical Containment'' claim is an empirical observation, not a formal proof (noted in Section 4.4).

\item {\bf Experimental result reproducibility}
    \item[] Answer: \answerYes{}
    \item[] Justification: Code and experimental configurations will be released upon acceptance to the conference, pending final public release approval from the Naval Surface Warfare Center.

\item {\bf Open access to data and code}
    \item[] Answer: \answerYes{}
    \item[] Justification: Code and experimental configurations will be released upon acceptance to the conference, pending final public release approval from the Naval Surface Warfare Center.

\item {\bf Experimental setting/details}
    \item[] Answer: \answerYes{}
    \item[] Justification: Section 4 specifies hardware (Quadro RTX 5000), probe sizes (50/100), fold counts (5), and all hyperparameters. Appendix~\ref{app:compute} provides computational budget.

\item {\bf Experiment statistical significance}
    \item[] Answer: \answerYes{}
    \item[] Justification: Table~\ref{tab:comparison} reports 95\% CIs and paired $t$-test $p$-values. We explicitly note where results are not significant (Llama $p{=}0.37$). Table~\ref{tab:s1} reports 95\% CIs via bootstrap.

\item {\bf Experiments compute resources}
    \item[] Answer: \answerYes{}
    \item[] Justification: Appendix~\ref{app:compute} provides a full breakdown of GPU-hours per experiment category, including failed/preliminary experiments ($\sim$210 total GPU-hours on a single Quadro RTX 5000).

\item {\bf Code of ethics}
    \item[] Answer: \answerYes{}
    \item[] Justification: Our research conforms to the NeurIPS Code of Ethics. We disclose the prompt injection vulnerability we discovered (Section~\ref{sec:injection}) to promote defensive research.

\item {\bf Broader impacts}
    \item[] Answer: \answerYes{}
    \item[] Justification: The Broader Impact section discusses dual-use risks of automated steering discovery and responsible disclosure of injection templates.

\item {\bf Safeguards}
    \item[] Answer: \answerYes{}
    \item[] Justification: Injection templates will be released with defensive countermeasure documentation. The steering discovery code does not enable capabilities beyond what is already possible with existing open-source steering tools.

\item {\bf Licenses for existing assets}
    \item[] Answer: \answerYes{}
    \item[] Justification: All models are used under their respective open-source licenses (Llama Community License, Apache 2.0 for OLMo and Gemma). Datasets are publicly available research datasets.

\item {\bf New assets}
    \item[] Answer: \answerYes{}
    \item[] Justification: Code and experimental configurations will be released upon acceptance to the conference, pending final public release approval from the Naval Surface Warfare Center.

\item {\bf Crowdsourcing and research with human subjects}
    \item[] Answer: \answerNA{}
    \item[] Justification: This research does not involve crowdsourcing or human subjects.

\item {\bf Institutional review board (IRB) approvals or equivalent for research with human subjects}
    \item[] Answer: \answerNA{}
    \item[] Justification: This research does not involve human subjects.

\item {\bf Declaration of LLM usage}
    \item[] Answer: \answerYes{}
    \item[] Justification: LLMs (Claude) were used to assist with code development and manuscript preparation. The core scientific methodology, experimental design, and analysis were performed by the authors. LLMs are also the subject of study in this work.

\end{enumerate}

\end{document}